\documentclass[a4paper]{llncs}

\usepackage[T1]{fontenc}
\usepackage[utf8]{inputenc}
\usepackage{graphicx}
\usepackage{amsmath}
\usepackage{booktabs}
\usepackage{multirow}
\usepackage{array}
\usepackage{tikz}
\usepackage{pgfplots}\pgfplotsset{compat=1.17}
\usetikzlibrary{positioning,arrows.meta,fit,backgrounds,calc}
\usepackage{xcolor}
\usepackage{xurl}
\usepackage{lmodern}
\usepackage{microtype}
\usepackage[hidelinks]{hyperref}

\newcolumntype{L}[1]{>{\raggedright\arraybackslash}p{#1}}

\usepackage{etoolbox}
\makeatletter
\patchcmd{\@maketitle}{\@title \par}{\vspace*{3mm}\@title \par\vspace{-3mm}}{}{\PackageError{rita}{Title spacing patch failed}{Use the supplied llncs class.}}
\makeatother

\begin{document}

\title{An Integrated Open-Source Stack for Vision-Based Autonomous Navigation of Wheeled Rovers}
\titlerunning{Open-Source Stack for Vision-Based Rover Navigation}

\author{Iaroslav Kolomiets\inst{1}\thanks{All six authors contributed equally to this work. Corresponding author: Iaroslav Kolomiets (rslvklmc@gmail.com).} \and
Iaroslav Sheipak\inst{1}$^{\star}$ \and
Mikhail Reppo\inst{1}$^{\star}$ \and
Vsevolod Egorov\inst{1}$^{\star}$ \and
Viktor Chertopolokhov\inst{2}$^{\star}$ \and
Isaac Chairez\inst{3}$^{\star}$}
\authorrunning{I. Kolomiets et al.}

\institute{MSU Rover Team\\
\url{https://msuroverteam.org/}\\
\email{rslvklmc@gmail.com, iaroslav.sheipak@gmail.com, sa.derfar23@gmail.com, game21rus@gmail.com}
\and
Research Institute for Brain Development and Peak Performance, RUDN University, 117198 Moscow, Russia\\
\email{psvr.msu@gmail.com}
\and
Institute of Advanced Materials for Sustainable Manufacturing, Tecnologico de Monterrey, Campus Guadalajara, Zapopan, Jalisco, Mexico\\
\email{isaac.chairez@tec.mx}}

\maketitle
\pagestyle{empty}\thispagestyle{empty}

\begin{abstract}
Vision-based rover autonomy requires integrating perception, localization, planning and platform-specific control under limited sensing and computing resources. We present an open-source ROS~2 stack for following directional arrows, reaching a terminal cone and returning autonomously to the start. It connects YOLOv8n detection and geometric direction, range and bearing estimation with RGB-D SLAM, wheel--inertial state estimation, kinematics-aware planning, behavior-tree mission management and STM32 actuation. Standard interfaces separate reusable autonomy from the hardware layer; a proportional path-following plugin supports low-speed rover motion. Public resources include software, configuration, a labeled dataset, trained weights and evaluation logs. On 478 held-out images, the detector achieves mAP@0.5 of 0.720 under the Ultralytics evaluator and outperforms a classical OpenCV baseline under a shared evaluation protocol. The reported perception latency is 25~ms on a Jetson Orin NX. Three missions, one indoor and two outdoor, demonstrate marker-guided navigation and autonomous return without operator motion commands or reported collisions. Field results document integration failures and corrective choices; repeated instrumented trials remain necessary to quantify physical navigation errors and reliability.

\end{abstract}

\section{Introduction}

Autonomous ground vehicles are central to exploration, agriculture, logistics, inspection and search-and-rescue \cite{nahavandi2025comprehensive}, and wheeled rovers balance mechanical simplicity, energy efficiency and terrain adaptability \cite{bhatnagar2024design}. Their key capability is autonomous navigation \cite{youssef2024autonomous}, and cameras, being lightweight, inexpensive and semantically rich, make vision-based navigation attractive for low-cost research and educational platforms \cite{chen2024integrated,woods2014seeker}.

Lighting, texture, sensor noise and on-board computing constraints complicate deployment, while reusable integration remains difficult~\cite{yang2023vision}. ROS~2 provides middleware and interface conventions~\cite{macenski2022ros2}; established components include OMPL~\cite{sucan2012ompl}, Nav2~\cite{macenski2020marathon}, RTAB-Map~\cite{labbe2019rtabmap}, integrated vehicle planners~\cite{darweesh2017open}, and AprilTag/ArUco fiducials~\cite{olson2011apriltag,garrido2014aruco}. Our pipeline instead interprets uncoded arrows and cones, converts observations into map-frame goals and executes them on a non-holonomic rover under mission-level supervision~ by a behavior tree\cite{colledanchise2018behavior}.

The contributions of this work are:
\begingroup
\renewcommand{\labelenumi}{(\roman{enumi})}
\begin{enumerate}
\item An integrated open-source ROS~2 stack from perception to STM32 actuation, separating reusable autonomy from platform-specific control.
\item A YOLOv8n marker-perception pipeline with geometric direction, range and bearing estimation, public data and weights, and a classical OpenCV comparison.
\item Behavior-tree marker search, goal generation and return, integrated with kinematics-aware planning and proportional path following.
\item Three documented physical-rover missions, including integration failures, corrective choices and per-mission outcomes.
\end{enumerate}
\endgroup
Table~\ref{tab:components} distinguishes developed and reused components.

\section{System Architecture}
\label{sec:arch}

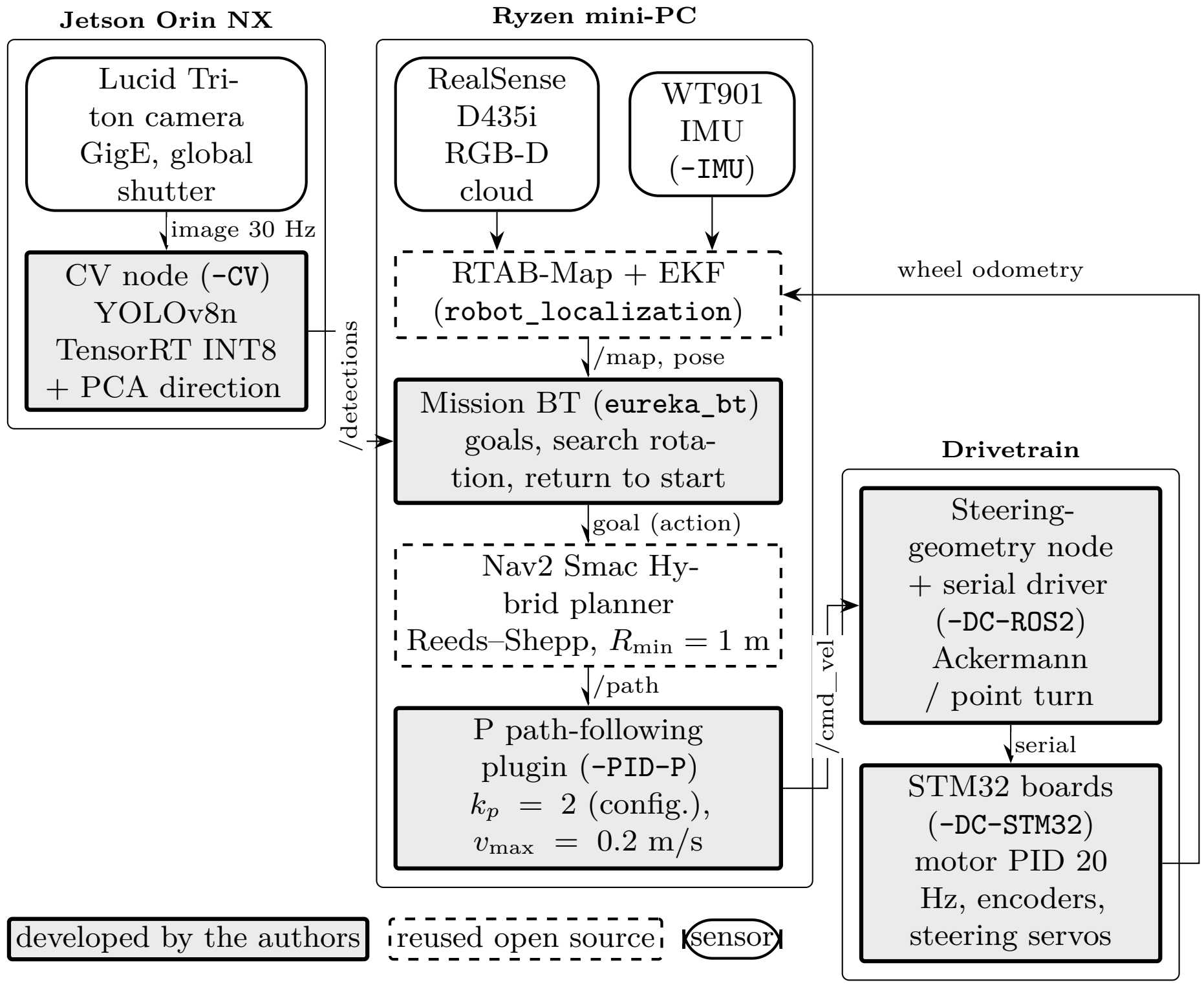
\begin{figure}[t]
\centering
\resizebox{\textwidth}{!}{%
\begin{tikzpicture}[
  font=\footnotesize,
  >={Stealth[length=2mm]},
  box/.style={align=center, inner sep=3pt, minimum height=8mm, rounded corners=1pt},
  dev/.style={box, draw, very thick, fill=black!8},
  reuse/.style={box, draw, thick, dashed, fill=white},
  sens/.style={box, draw, thick, fill=white, rounded corners=3mm, minimum height=7mm},
  host/.style={draw, thin, rounded corners=2pt, inner sep=5pt},
  lbl/.style={font=\scriptsize, fill=white, inner sep=1pt},
  node distance=4mm and 9mm]
\node[sens, text width=27mm] (cam) {Lucid Triton camera\\GigE, global shutter};
\node[dev, text width=27mm, below=of cam] (cv) {CV node (\texttt{-CV})\\YOLOv8n TensorRT INT8\\+ PCA direction};
\node[sens, text width=19mm, right=of cam] (rs) {RealSense D435i\\RGB-D cloud};
\node[sens, text width=15mm, right=3mm of rs] (imu) {WT901 IMU\\(\texttt{-IMU})};
\node[reuse, text width=38mm, below=of rs.south west, anchor=north west] (slam) {RTAB-Map + EKF\\(\texttt{robot\_\allowbreak localization})};
\node[dev, text width=38mm, below=of slam] (bt) {Mission BT (\texttt{eureka\_bt})\\goals, search rotation, return to start};
\node[reuse, text width=38mm, below=of bt] (smac) {Nav2 Smac Hybrid planner\\Reeds--Shepp, $R_{\min}=1$ m};
\node[dev, text width=38mm, below=of smac] (ctl) {P path-following plugin (\texttt{-PID-P})\\$k_p=2$ (config.), $v_{\max}=0.2$ m/s};
\node[dev, text width=29mm, right=8mm of smac] (steer) {Steering-geometry node\\+ serial driver (\texttt{-DC-ROS2})\\Ackermann / point turn};
\node[dev, text width=29mm, below=of steer] (stm) {STM32 boards (\texttt{-DC-STM32})\\motor PID 20 Hz, encoders, steering servos};
\begin{scope}[on background layer]
\node[host, fit=(cam)(cv), label={[font=\scriptsize\bfseries]above:Jetson Orin NX}] (h1) {};
\node[host, fit=(rs)(imu)(slam)(bt)(smac)(ctl), label={[font=\scriptsize\bfseries]above:Ryzen mini-PC}] (h2) {};
\node[host, fit=(steer)(stm), label={[font=\scriptsize\bfseries]above:Drivetrain}] (h3) {};
\end{scope}
\draw[->] (cam) -- (cv) node[lbl, midway, right] {image 30 Hz};
\coordinate (gab) at ($(h1.east)!0.5!(h2.west)$);
\draw[->] (cv.east) -- (cv.east -| gab) |- (bt.west);
\node[lbl, rotate=90] at ($(cv.east -| gab)!0.5!(bt.west -| gab)$) {/detections};
\draw[->] (rs) -- (rs |- slam.north);
\draw[->] (imu) -- (imu |- slam.north);
\draw[->] (slam) -- (bt) node[lbl, midway, right] {/map, pose};
\draw[->] (bt) -- (smac) node[lbl, midway, right] {goal (action)};
\draw[->] (smac) -- (ctl) node[lbl, midway, right] {/path};
\coordinate (gbc) at ($(h2.east)!0.5!(h3.west)$);
\draw[->] (ctl.east) -- (ctl.east -| gbc) |- (steer.west);
\node[lbl, rotate=90] at ($(ctl.east -| gbc)!0.5!(steer.west -| gbc)$) {/cmd\_vel};
\draw[->] (steer) -- (stm) node[lbl, midway, right] {serial};
\coordinate (odx) at ($(h3.east)+(2mm,0)$);
\draw[->] (stm.east) -- (stm.east -| odx) |- (slam.east);
\node[lbl] at ($($(slam.east)!0.5!(slam.east -| odx)$)+(0,2.5mm)$) {wheel odometry};
\coordinate (lg) at ($(h1.west |- h2.south)+(0,-3mm)$);
\node[dev, minimum height=4mm, inner sep=2pt, anchor=north west] (l1) at (lg) {developed by the authors};
\node[reuse, minimum height=4mm, inner sep=2pt, right=2mm of l1] (l2) {reused open source};
\node[sens, minimum height=4mm, inner sep=2pt, right=2mm of l2] (l3) {sensor};
\end{tikzpicture}}
\caption{Deployed perception-to-action stack. Boxes with a gray fill were developed by the authors; dashed boxes are reused open-source components; rounded boxes are sensors. The path-following controller is a proportional (P) controller; its displayed gain is from the inspected configuration (Sect.~\ref{sec:control}).}
\label{fig:arch}
\end{figure}

Three open-source libraries\footnote{\url{https://github.com/KodII-rover}; Table~\ref{tab:components} abbreviates repository names by their suffix after \texttt{MSUROVERTEAM}.} integrate hardware control, navigation and vision (Fig.~\ref{fig:arch}). Table~\ref{tab:components} identifies developed/reused components and configuration files. The documented environment is Ubuntu~22.04, ROS~2 Humble, Humble binaries of Nav2, RTAB-Map (\texttt{rtabmap\_ros}) and \texttt{robot\_localization}, BehaviorTree.CPP and YOLOv8/TensorRT. Portability follows the interface design but has not been demonstrated on a second rover.

\begin{table}[t]
\caption{Components of the deployed stack, their origin and where they are configured.}
\label{tab:components}
\centering\scriptsize
\setlength{\tabcolsep}{4pt}
\begin{tabular}{L{1.5cm}L{4.3cm}L{1.6cm}L{3.6cm}}
\toprule
\textbf{Layer} & \textbf{Component} & \textbf{Origin} & \textbf{Repository / file} \\
\midrule
Actuation & STM32 firmware: motor PID/STC loops, steering position control, encoders & developed & \texttt{-DC-STM32} \\
HAL & ROS~2 serial driver and steering-geometry node (Ackermann, point turn, strafe) & developed & \texttt{-DC-ROS2/}\newline\texttt{eureka\_movement\_2} \\
Sensing & IMU driver (WitMotion WT901) & developed & \texttt{-IMU} \\
Localization & Wheel odometry + IMU EKF; RGB-D SLAM & reused & \texttt{robot\_localization}, RTAB-Map;\newline\texttt{-SLAM/}\newline\texttt{eureka\_localization/}\newline\texttt{config/} \\
Mapping & Occupancy grid from depth cloud & reused & RTAB-Map \texttt{Grid/*} params in\newline\texttt{-SLAM/.../}\newline\texttt{nav2tune.launch.py} \\
Planning & Smac Hybrid-A* planner, costmaps, recoveries & reused & Nav2; \texttt{-SLAM/}\newline\texttt{eureka\_navigation/}\newline\texttt{config/nav2.yaml} \\
Control & P path-following controller (Nav2 plugin) & developed & \texttt{-PID-P}; gains in \texttt{nav2.yaml} \\
Mission & Behavior tree, goal generation, search rotation, return home; C++ port of \texttt{simple\_commander} & developed & \texttt{-SLAM/eureka\_bt};\newline params in\newline\texttt{strategy.launch.py} \\
Perception & YOLOv8n detector, PCA direction, range/bearing tables, ROS~2 node & developed (model: Ultralytics) & \texttt{-CV} \\
\bottomrule
\end{tabular}
\end{table}

\paragraph{Software provenance.}
Repository revisions and reported software versions document the released stack; exact deployment revisions and package versions for the 2025 trials were not archived~\cite{projectreport}.


\subsection{Hardware Abstraction and Drivetrain}
\label{sec:hal}

The platform-specific layer consists of STM32-based boards and a Python ROS~2 driver. Two 24~V DC motors with 100:1 gearboxes per side drive the wheels; their speed is measured by magnetic encoders (34 ticks per motor revolution) and closed at 20~Hz by PID or super-twisting loops. Each wheel is steered by a NEMA~17 stepper (50:1 gearbox) position-controlled from an AS5600 absolute encoder, so missed steps on rough terrain do not accumulate. The driver talks to the boards over a checksummed serial protocol and works with a steering-geometry node that maps the standard \texttt{cmd\_vel} interface onto four steering angles and four wheel speeds (wheelbase 0.76~m, track 0.94~m, wheel diameter 0.30~m). The angular component is interpreted as path curvature and clamped to $\pm1.4$~m$^{-1}$, i.e. a kinematic turning radius $\geq 0.71$~m; the steering angles for a commanded radius come from a precomputed lookup table, and the drive speeds of the four wheels are scaled by the current measured steering angles so that mode transitions occur with the wheels already re-aligned. Besides Ackermann steering the drivetrain offers strafing (teleoperation only) and a \emph{point turn}: the wheels are steered to $\pm39^\circ$, tangent to the circle through the four contact points, and the two sides are driven in opposite directions, so the rover rotates about its vertical axis. The point turn is a distinct steering mode of the independently steered drivetrain, not a zero-radius Ackermann maneuver; the mission layer requests it with a sentinel angular command ($|\omega|>90$) and uses it only for the stationary marker search (Sect.~\ref{sec:bt}). Path following always uses the Ackermann mode.

\subsection{Navigation}
\label{sec:nav}

\subsubsection{Data flow.}
The CV node on the Jetson publishes marker detections (class, arrow direction, distance, bearing, confidence) at 30~Hz. On the navigation computer, wheel odometry (20~Hz) and IMU orientation are fused by an extended Kalman filter \cite{moore2016robotlocalization}; RTAB-Map \cite{labbe2019rtabmap} consumes the RGB-D stream of an Intel RealSense D435i and this odometry, maintains the pose graph with loop closure and publishes the map and the \texttt{map}$\to$\texttt{odom} correction. The BT converts detections into map-frame goals and sends them to Nav2 \cite{macenski2020marathon} as actions; the Smac Hybrid planner produces an Ackermann-feasible path, and the P controller plugin turns it into \texttt{cmd\_vel} commands for the HAL. The navigation side is started by \texttt{nav2tune.launch.py} (sensors, EKF, RTAB-Map, Nav2 with \texttt{nav2.yaml}) and \texttt{strategy.launch.py} (BT).

\subsubsection{Localization and mapping.}
RTAB-Map uses 10~Hz detection, GTSAM optimization with robust kernels, and node insertion every 0.05~m or 0.05~rad. Its 0.1~m occupancy grid uses D435i returns between \texttt{Grid/RangeMin}=0.05~m and \texttt{Grid/RangeMax}=4.0~m. The report specifies a 3~m corrective cutoff~\cite{projectreport}; the exact deployed threshold is unverified. Filtering addressed invalid long-range returns rasterized as occupied beyond about 3~m, blocking distant goals (Sect.~\ref{sec:field}). Discarded returns leave cells unobserved. Nav2 uses the grid as a static layer with inflation radius 0.5~m and cost scaling 10; \texttt{track\_unknown\_space} is disabled and \texttt{allow\_unknown} enabled. Thus goals up to the 10~m visual range can traverse unobserved space; obstacles are incorporated within the reliable depth range of about 3--4~m (costmap updates 4~Hz, replanning 1~Hz). Local/global rolling costmaps span 10/20~m, with 0.15~m cells and a 1.2$\times$1.1~m footprint. Loop closure corrects drift, but its contribution to return accuracy was not isolated experimentally.

\subsubsection{Path planning and control.}
\label{sec:control}
The Smac Hybrid planner~\cite{macenski2026cost} is configured with the Reeds--Shepp motion model, 64 heading bins, a minimum turning radius of 1.0~m (conservative with respect to the 0.71~m kinematic limit), goal tolerance 0.2~m, reverse and direction-change penalties 1.9, non-straight penalty 1.1, cost penalty 3.3 and a 3.5~s planning budget. In simulation and field tests at 0.2~m/s, the tested MPPI~\cite{williams2018information} configuration produced frequent re-steering and back-and-forth motion on rough ground; tight velocity bounds restrict sampled trajectories. We implemented a proportional Nav2 path-following plugin, related to pure pursuit~\cite{macenski2023regulated}. At 10~Hz it selects four path poses from the closest pose, places a look-ahead point $L=1.0$~m along the rover heading, and computes its signed lateral offset $e_y$ from the segment line and heading error $e_\theta$ to the segment end. The command is
\begin{equation}
\omega = -k_p\, e_y, \qquad v = v_{\max}\cos e_\theta,
\end{equation}
with $k_p = 2.0$, $v_{\max}=0.2$~m/s, and the sign of $v$ reversed when the segment heading differs from the rover heading by more than $90^\circ$ (Reeds--Shepp reversing). Integral and derivative terms are implemented but set to zero. The inspected \texttt{nav2.yaml} sets $k_p=2.0$, whereas the controller README gives $K_p=1$; the gain used during the reported trials was not archived. A velocity smoother limits the output to 0.41~m/s and 0.7~rad/s, and the goal checker accepts a pose within 0.4~m and 0.5~rad. The preference for the P controller over MPPI holds for this low-speed configuration only; a quantitative controller comparison was not archived (Sect.~\ref{sec:sim}).

\subsubsection{Behavior tree and object-based navigation.}
\label{sec:bt}
The BT~\cite{colledanchise2018behavior} (BehaviorTree.CPP, \texttt{eureka\_bt}, a sequence of five action nodes ticked at 10~Hz) implements the mission loop (search a marker, go to it, stop, turn and search again; on the cone, return home). Detections are filtered over a buffer of ten messages, requiring agreement within 2.0~m and $3^\circ$ before a marker is accepted; detections beyond 11~m are ignored. For an accepted marker at distance $d$ and bearing $\beta$, the goal is placed in the map frame at $d-2.1$~m along $\beta$ from the current pose, i.e. inside the 2~m stop zone, and sent through a NavigateToPose action; if the rover comes within 5~m of the goal the goal is republished from the refreshed estimate, and a goal that is not reached within 60~s is re-issued. World-frame goals let the rover continue toward a marker that has left the field of view. On arrival it stops, pre-steers the wheels into the point-turn configuration (5~s), and rotates in the direction indicated by the arrow until the next marker is detected within $\pm9^\circ$; the rotation is executed open-loop by the BT, outside Nav2 and hence without costmap collision checking. A collision-free approach does not guarantee a collision-free swept footprint during rotation; checking that footprint against the costmap before turning is future work. Nav2's own replanning-and-recovery tree (costmap clearing, backing up, waiting) handles planning failures during travel. After the cone is reached the rover backs up 0.15~m and navigates to the stored start pose. Because topic-level commands to Nav2 gave no preemption or completion feedback, the Python \texttt{simple\_commander} was ported to C++ so the BT drives Nav2 through action clients with cancellation and progress monitoring.

\subsection{Machine-Learning Perception}
\label{sec:perc}

A Lucid Triton TRI016S-CC camera (Sony IMX273 global shutter, GigE Vision, Arena SDK) delivers 640$\times$480 frames at 30~Hz to the Jetson Orin NX. The detector is YOLOv8 nano \cite{redmon2016yolo,ultralytics2023yolov8} (3.2~M parameters) with two classes, navigation arrow and traffic cone. The directional marker is a 300$\times$200~mm white plate with a black arrow mounted 100--150~cm above the ground; the terminal target is an orange traffic cone per GOST~32758-2014.

\subsubsection{Dataset and training.}
The public dataset\footnote{\url{https://www.kaggle.com/datasets/prttyflvck00/msuroverteam-cv-dataset}} contains 6{,}147 images with 20{,}303 instances (5{,}145 arrows, 15{,}158 cones): rover field footage plus stock photographs, rendered arrows, Mars-analog footage and public cone-course videos that widen backgrounds, scales and lighting. Splits are 4{,}298 / 1{,}371 / 478 images (train/val/test); frames of one video stay in one split and near-duplicates were removed. Arrow direction is not a detector class, so flips and affine augmentations remain label-safe. Training used SGD (momentum 0.937, lr 0.01, weight decay $5\times10^{-4}$), mosaic/HSV/flip/erasing/affine augmentation, 640$\times$640 input and early stopping (patience 50) at epoch 197; the epoch-147 checkpoint was deployed. The evaluated weights and logs are published on Hugging Face\footnote{\url{https://huggingface.co/StrongUzumaki/rover-navigation-stack-yolov8n}; evaluated release; \texttt{MSUROVERTEAM-CV} instead bundles older 2024 weights.}.

\subsubsection{Detection performance.}
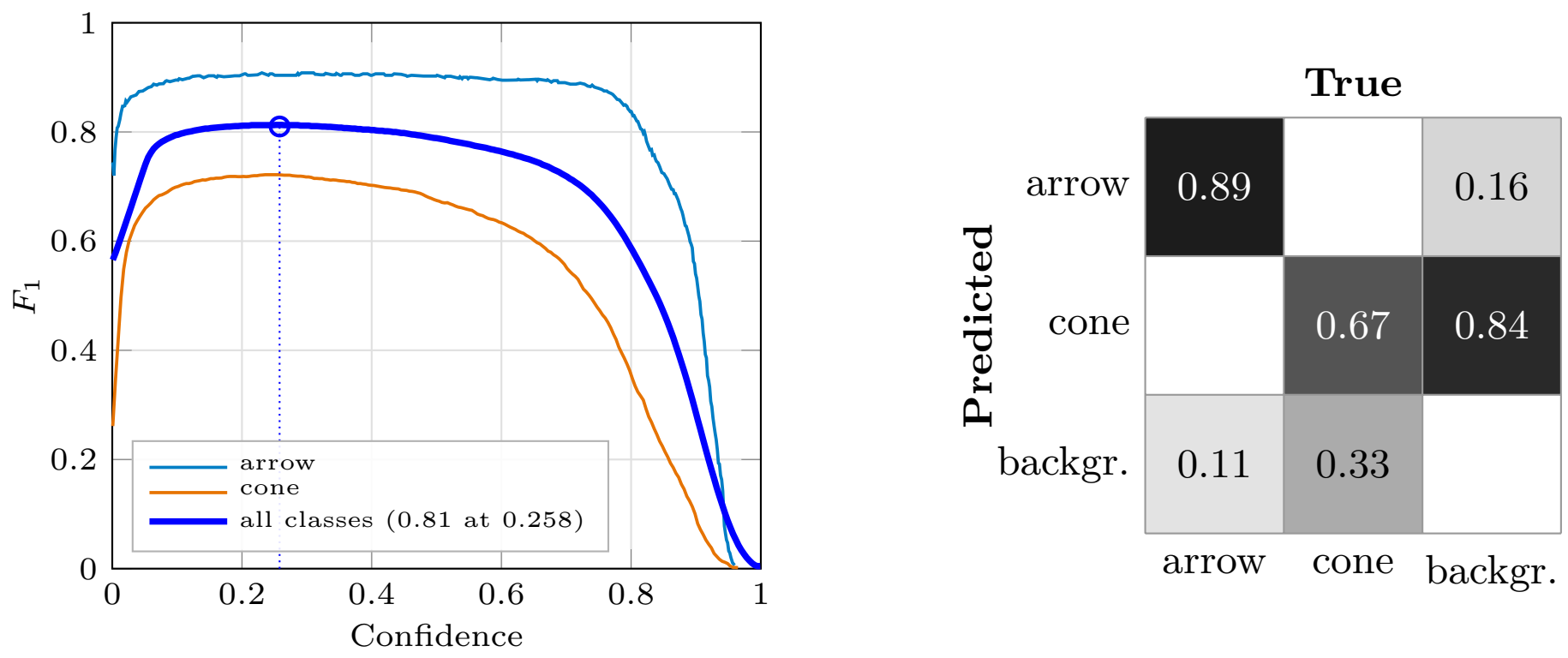
\begin{figure}[t]
\centering
\begin{minipage}[c]{0.50\textwidth}\centering
\begin{tikzpicture}
\begin{axis}[width=1.08\linewidth,height=0.95\linewidth,font=\scriptsize,
  xlabel={Confidence},ylabel={$F_1$},xmin=0,xmax=1,ymin=0,ymax=1,
  xtick={0,0.2,...,1},ytick={0,0.2,...,1},
  xlabel style={yshift=2pt},ylabel style={yshift=-4pt},
  grid=major,grid style={black!12},tick label style={/pgf/number format/fixed},
  legend style={at={(0.03,0.03)},anchor=south west,font=\tiny,draw=black!30,fill=white,fill opacity=0.9,text opacity=1,row sep=-2pt},
  legend cell align=left,clip=false]
\addplot[cyan!60!blue,line width=0.7pt] coordinates {
(0.0008,0.7442)(0.0025,0.7200)(0.0042,0.7668)(0.0059,0.7897)(0.0076,0.8076)(0.0092,0.8103)
(0.0117,0.8228)(0.0134,0.8341)(0.0159,0.8471)(0.0176,0.8471)(0.0193,0.8480)(0.0210,0.8570)
(0.0227,0.8543)(0.0260,0.8597)(0.0285,0.8642)(0.0327,0.8660)(0.0369,0.8678)(0.0411,0.8741)
(0.0470,0.8750)(0.0554,0.8813)(0.0604,0.8831)(0.0638,0.8876)(0.0688,0.8858)(0.0721,0.8880)
(0.0789,0.8885)(0.0822,0.8930)(0.0839,0.8939)(0.0889,0.8921)(0.0982,0.8952)(0.1015,0.8948)
(0.1032,0.8975)(0.1082,0.8957)(0.1166,0.8984)(0.1183,0.8984)(0.1242,0.9020)(0.1258,0.9020)
(0.1317,0.9029)(0.1351,0.9029)(0.1409,0.8993)(0.1485,0.9029)(0.1535,0.9029)(0.1569,0.9029)
(0.1586,0.9015)(0.1611,0.9002)(0.1628,0.9029)(0.1720,0.9011)(0.1820,0.9020)(0.1888,0.9029)
(0.1946,0.9029)(0.1963,0.9029)(0.2039,0.9065)(0.2055,0.9065)(0.2089,0.9038)(0.2122,0.9038)
(0.2240,0.9047)(0.2307,0.9083)(0.2408,0.9038)(0.2433,0.9065)(0.2450,0.9065)(0.2466,0.9065)
(0.2508,0.9038)(0.2525,0.9038)(0.2836,0.9038)(0.2861,0.9074)(0.2878,0.9047)(0.2919,0.9083)
(0.2936,0.9083)(0.2953,0.9083)(0.2970,0.9083)(0.2987,0.9083)(0.3020,0.9083)(0.3037,0.9083)
(0.3104,0.9083)(0.3129,0.9056)(0.3146,0.9056)(0.3213,0.9047)(0.3230,0.9047)(0.3263,0.9047)
(0.3289,0.9074)(0.3389,0.9047)(0.3440,0.9047)(0.3456,0.9074)(0.3515,0.9038)(0.3532,0.9038)
(0.3557,0.9065)(0.3691,0.9065)(0.3742,0.9038)(0.3758,0.9038)(0.3775,0.9038)(0.3792,0.9038)
(0.3809,0.9038)(0.3867,0.9065)(0.3884,0.9065)(0.3935,0.9065)(0.3985,0.9038)(0.4002,0.9038)
(0.4060,0.9065)(0.4169,0.9065)(0.4186,0.9065)(0.4211,0.9038)(0.4295,0.9038)(0.4312,0.9038)
(0.4346,0.9065)(0.4455,0.9056)(0.4530,0.9033)(0.4547,0.9020)(0.4614,0.9038)(0.4799,0.9024)
(0.4815,0.9024)(0.4891,0.9024)(0.4908,0.9024)(0.4992,0.9024)(0.5008,0.9038)(0.5126,0.8993)
(0.5143,0.9020)(0.5168,0.9020)(0.5185,0.9020)(0.5201,0.9020)(0.5218,0.9020)(0.5344,0.8984)
(0.5361,0.9011)(0.5394,0.8975)(0.5411,0.9002)(0.5470,0.9002)(0.5487,0.8975)(0.5512,0.9002)
(0.5529,0.9002)(0.5604,0.8993)(0.5621,0.8993)(0.5822,0.8957)(0.5856,0.8970)(0.5872,0.8984)
(0.5940,0.8961)(0.6015,0.8948)(0.6082,0.8948)(0.6242,0.8952)(0.6258,0.8966)(0.6292,0.8952)
(0.6401,0.8957)(0.6418,0.8957)(0.6535,0.8966)(0.6552,0.8939)(0.6569,0.8966)(0.6695,0.8943)
(0.6711,0.8957)(0.6770,0.8912)(0.6804,0.8939)(0.6862,0.8939)(0.6879,0.8903)(0.6896,0.8903)
(0.6913,0.8903)(0.6930,0.8903)(0.6980,0.8903)(0.7064,0.8876)(0.7114,0.8876)(0.7148,0.8876)
(0.7215,0.8894)(0.7248,0.8885)(0.7282,0.8876)(0.7332,0.8840)(0.7374,0.8831)(0.7399,0.8858)
(0.7433,0.8822)(0.7508,0.8795)(0.7550,0.8777)(0.7584,0.8750)(0.7626,0.8750)(0.7643,0.8732)
(0.7659,0.8723)(0.7693,0.8705)(0.7718,0.8687)(0.7760,0.8633)(0.7785,0.8597)(0.7819,0.8588)
(0.7869,0.8516)(0.7886,0.8489)(0.7911,0.8498)(0.7936,0.8471)(0.7953,0.8422)(0.7970,0.8399)
(0.7987,0.8372)(0.8003,0.8372)(0.8029,0.8296)(0.8045,0.8300)(0.8087,0.8201)(0.8104,0.8192)
(0.8146,0.8058)(0.8171,0.8040)(0.8188,0.7977)(0.8205,0.8004)(0.8221,0.7932)(0.8238,0.7830)
(0.8263,0.7752)(0.8289,0.7671)(0.8314,0.7680)(0.8331,0.7572)(0.8356,0.7581)(0.8372,0.7513)
(0.8389,0.7464)(0.8414,0.7437)(0.8440,0.7365)(0.8473,0.7257)(0.8507,0.7235)(0.8523,0.7203)
(0.8557,0.7140)(0.8591,0.7037)(0.8624,0.6951)(0.8649,0.6942)(0.8666,0.6871)(0.8683,0.6871)
(0.8725,0.6754)(0.8750,0.6655)(0.8767,0.6592)(0.8784,0.6544)(0.8800,0.6470)(0.8817,0.6376)
(0.8834,0.6286)(0.8851,0.6271)(0.8893,0.6088)(0.8909,0.5992)(0.8926,0.5944)(0.8943,0.5877)
(0.8960,0.5627)(0.8977,0.5589)(0.8993,0.5400)(0.9010,0.5324)(0.9027,0.5166)(0.9044,0.4996)
(0.9060,0.4896)(0.9077,0.4654)(0.9094,0.4573)(0.9111,0.4317)(0.9128,0.4159)(0.9144,0.3925)
(0.9161,0.3862)(0.9178,0.3533)(0.9195,0.3450)(0.9211,0.3327)(0.9228,0.3064)(0.9245,0.2968)
(0.9262,0.2765)(0.9279,0.2617)(0.9295,0.2486)(0.9312,0.2363)(0.9329,0.2181)(0.9346,0.1933)
(0.9362,0.1915)(0.9379,0.1708)(0.9396,0.1632)(0.9413,0.1421)(0.9438,0.0782)(0.9455,0.0643)
(0.9471,0.0526)(0.9488,0.0474)(0.9505,0.0324)(0.9522,0.0288)(0.9539,0.0202)(0.9564,0.0108)
(0.9589,0.0072)(0.9597,0.0063)};
\addlegendentry{arrow}
\addplot[orange!90!black,line width=0.7pt] coordinates {
(0.0008,0.2612)(0.0050,0.3462)(0.0092,0.4204)(0.0134,0.4960)(0.0176,0.5486)(0.0218,0.5796)
(0.0260,0.6012)(0.0302,0.6151)(0.0344,0.6286)(0.0386,0.6376)(0.0428,0.6457)(0.0470,0.6533)
(0.0512,0.6605)(0.0554,0.6650)(0.0596,0.6700)(0.0638,0.6758)(0.0680,0.6799)(0.0721,0.6835)
(0.0763,0.6848)(0.0805,0.6871)(0.0847,0.6897)(0.0889,0.6933)(0.0931,0.6965)(0.0973,0.6987)
(0.1015,0.7001)(0.1057,0.7023)(0.1099,0.7041)(0.1141,0.7055)(0.1183,0.7055)(0.1225,0.7068)
(0.1267,0.7091)(0.1309,0.7100)(0.1351,0.7109)(0.1393,0.7122)(0.1435,0.7118)(0.1477,0.7131)
(0.1518,0.7145)(0.1560,0.7145)(0.1602,0.7154)(0.1644,0.7154)(0.1686,0.7158)(0.1728,0.7172)
(0.1770,0.7181)(0.1812,0.7185)(0.1854,0.7185)(0.1896,0.7185)(0.1938,0.7181)(0.1980,0.7181)
(0.2022,0.7185)(0.2064,0.7181)(0.2106,0.7185)(0.2148,0.7194)(0.2190,0.7199)(0.2232,0.7199)
(0.2273,0.7208)(0.2315,0.7212)(0.2357,0.7217)(0.2399,0.7217)(0.2441,0.7217)(0.2483,0.7217)
(0.2525,0.7217)(0.2567,0.7217)(0.2609,0.7212)(0.2651,0.7208)(0.2693,0.7208)(0.2735,0.7208)
(0.2777,0.7199)(0.2819,0.7194)(0.2861,0.7190)(0.2903,0.7190)(0.2945,0.7185)(0.2987,0.7176)
(0.3029,0.7176)(0.3070,0.7167)(0.3112,0.7149)(0.3154,0.7149)(0.3196,0.7145)(0.3238,0.7145)
(0.3280,0.7131)(0.3322,0.7127)(0.3364,0.7118)(0.3406,0.7122)(0.3448,0.7118)(0.3490,0.7109)
(0.3532,0.7104)(0.3574,0.7100)(0.3616,0.7091)(0.3658,0.7091)(0.3700,0.7077)(0.3742,0.7073)
(0.3784,0.7055)(0.3826,0.7055)(0.3867,0.7046)(0.3909,0.7041)(0.3951,0.7028)(0.3993,0.7023)
(0.4035,0.7019)(0.4077,0.7010)(0.4119,0.7001)(0.4161,0.6992)(0.4203,0.6983)(0.4245,0.6978)
(0.4287,0.6974)(0.4329,0.6965)(0.4371,0.6960)(0.4413,0.6956)(0.4455,0.6938)(0.4497,0.6933)
(0.4539,0.6929)(0.4581,0.6915)(0.4622,0.6911)(0.4664,0.6902)(0.4706,0.6888)(0.4748,0.6875)
(0.4790,0.6853)(0.4832,0.6826)(0.4874,0.6803)(0.4916,0.6785)(0.4958,0.6763)(0.5000,0.6745)
(0.5042,0.6731)(0.5084,0.6709)(0.5126,0.6700)(0.5168,0.6704)(0.5210,0.6682)(0.5252,0.6664)
(0.5294,0.6641)(0.5336,0.6623)(0.5378,0.6614)(0.5419,0.6601)(0.5461,0.6587)(0.5503,0.6574)
(0.5545,0.6565)(0.5587,0.6533)(0.5629,0.6497)(0.5671,0.6479)(0.5713,0.6466)(0.5755,0.6439)
(0.5797,0.6416)(0.5839,0.6403)(0.5881,0.6394)(0.5923,0.6376)(0.5965,0.6349)(0.6007,0.6335)
(0.6049,0.6313)(0.6091,0.6286)(0.6133,0.6259)(0.6174,0.6237)(0.6216,0.6214)(0.6258,0.6183)
(0.6300,0.6160)(0.6342,0.6142)(0.6384,0.6106)(0.6426,0.6079)(0.6468,0.6048)(0.6510,0.6007)
(0.6552,0.5971)(0.6594,0.5949)(0.6636,0.5908)(0.6678,0.5868)(0.6720,0.5818)(0.6762,0.5796)
(0.6804,0.5751)(0.6846,0.5719)(0.6888,0.5679)(0.6930,0.5629)(0.6971,0.5576)(0.7013,0.5499)
(0.7055,0.5445)(0.7097,0.5396)(0.7139,0.5355)(0.7181,0.5301)(0.7223,0.5211)(0.7265,0.5121)
(0.7307,0.5072)(0.7349,0.5013)(0.7391,0.4946)(0.7433,0.4870)(0.7475,0.4793)(0.7517,0.4726)
(0.7559,0.4649)(0.7601,0.4591)(0.7643,0.4510)(0.7685,0.4415)(0.7727,0.4299)(0.7768,0.4200)
(0.7810,0.4060)(0.7852,0.3957)(0.7894,0.3858)(0.7936,0.3750)(0.7978,0.3629)(0.8020,0.3480)
(0.8062,0.3345)(0.8104,0.3237)(0.8146,0.3165)(0.8188,0.3089)(0.8230,0.2927)(0.8272,0.2783)
(0.8314,0.2684)(0.8356,0.2563)(0.8398,0.2455)(0.8440,0.2334)(0.8482,0.2244)(0.8523,0.2140)
(0.8565,0.2046)(0.8607,0.1933)(0.8649,0.1853)(0.8691,0.1749)(0.8733,0.1673)(0.8775,0.1547)
(0.8817,0.1457)(0.8859,0.1326)(0.8901,0.1241)(0.8943,0.1138)(0.8985,0.1007)(0.9027,0.0823)
(0.9069,0.0715)(0.9111,0.0625)(0.9153,0.0526)(0.9195,0.0450)(0.9237,0.0382)(0.9279,0.0279)
(0.9320,0.0225)(0.9362,0.0171)(0.9404,0.0135)(0.9446,0.0117)(0.9488,0.0094)(0.9530,0.0045)
(0.9572,0.0022)(0.9614,0.0022)(0.9648,0.0018)};
\addlegendentry{cone}
\addplot[blue,line width=1.4pt] coordinates {
(0.0008,0.5656)(0.0050,0.5782)(0.0092,0.5926)(0.0134,0.6070)(0.0176,0.6219)(0.0218,0.6367)
(0.0260,0.6515)(0.0302,0.6664)(0.0344,0.6817)(0.0386,0.6965)(0.0428,0.7118)(0.0470,0.7266)
(0.0512,0.7410)(0.0554,0.7536)(0.0596,0.7621)(0.0638,0.7693)(0.0680,0.7743)(0.0721,0.7783)
(0.0763,0.7819)(0.0805,0.7846)(0.0847,0.7873)(0.0889,0.7896)(0.0931,0.7918)(0.0973,0.7941)
(0.1015,0.7950)(0.1057,0.7968)(0.1099,0.7981)(0.1141,0.7995)(0.1183,0.8004)(0.1225,0.8013)
(0.1267,0.8026)(0.1309,0.8035)(0.1351,0.8044)(0.1393,0.8053)(0.1435,0.8062)(0.1477,0.8062)
(0.1518,0.8071)(0.1560,0.8076)(0.1602,0.8080)(0.1644,0.8085)(0.1686,0.8089)(0.1728,0.8094)
(0.1770,0.8098)(0.1812,0.8098)(0.1854,0.8103)(0.1896,0.8107)(0.1938,0.8107)(0.1980,0.8112)
(0.2022,0.8116)(0.2064,0.8116)(0.2106,0.8116)(0.2148,0.8125)(0.2190,0.8125)(0.2232,0.8125)
(0.2273,0.8125)(0.2315,0.8125)(0.2357,0.8125)(0.2399,0.8125)(0.2441,0.8125)(0.2483,0.8125)
(0.2525,0.8129)(0.2567,0.8129)(0.2609,0.8129)(0.2651,0.8125)(0.2693,0.8125)(0.2735,0.8125)
(0.2777,0.8125)(0.2819,0.8125)(0.2861,0.8125)(0.2903,0.8116)(0.2945,0.8116)(0.2987,0.8116)
(0.3029,0.8116)(0.3070,0.8107)(0.3112,0.8107)(0.3154,0.8107)(0.3196,0.8103)(0.3238,0.8098)
(0.3280,0.8098)(0.3322,0.8094)(0.3364,0.8094)(0.3406,0.8089)(0.3448,0.8085)(0.3490,0.8080)
(0.3532,0.8080)(0.3574,0.8071)(0.3616,0.8071)(0.3658,0.8067)(0.3700,0.8062)(0.3742,0.8062)
(0.3784,0.8053)(0.3826,0.8053)(0.3867,0.8044)(0.3909,0.8044)(0.3951,0.8044)(0.3993,0.8035)
(0.4035,0.8035)(0.4077,0.8026)(0.4119,0.8022)(0.4161,0.8022)(0.4203,0.8013)(0.4245,0.8013)
(0.4287,0.8008)(0.4329,0.7999)(0.4371,0.7999)(0.4413,0.7990)(0.4455,0.7981)(0.4497,0.7981)
(0.4539,0.7972)(0.4581,0.7963)(0.4622,0.7954)(0.4664,0.7950)(0.4706,0.7941)(0.4748,0.7936)
(0.4790,0.7927)(0.4832,0.7918)(0.4874,0.7909)(0.4916,0.7900)(0.4958,0.7891)(0.5000,0.7887)
(0.5042,0.7878)(0.5084,0.7869)(0.5126,0.7860)(0.5168,0.7851)(0.5210,0.7837)(0.5252,0.7828)
(0.5294,0.7819)(0.5336,0.7810)(0.5378,0.7801)(0.5419,0.7792)(0.5461,0.7783)(0.5503,0.7774)
(0.5545,0.7761)(0.5587,0.7747)(0.5629,0.7738)(0.5671,0.7729)(0.5713,0.7720)(0.5755,0.7711)
(0.5797,0.7698)(0.5839,0.7684)(0.5881,0.7675)(0.5923,0.7662)(0.5965,0.7653)(0.6007,0.7639)
(0.6049,0.7626)(0.6091,0.7612)(0.6133,0.7599)(0.6174,0.7590)(0.6216,0.7572)(0.6258,0.7558)
(0.6300,0.7545)(0.6342,0.7527)(0.6384,0.7513)(0.6426,0.7500)(0.6468,0.7482)(0.6510,0.7464)
(0.6552,0.7446)(0.6594,0.7428)(0.6636,0.7401)(0.6678,0.7383)(0.6720,0.7365)(0.6762,0.7338)
(0.6804,0.7311)(0.6846,0.7289)(0.6888,0.7266)(0.6930,0.7235)(0.6971,0.7208)(0.7013,0.7181)
(0.7055,0.7145)(0.7097,0.7118)(0.7139,0.7082)(0.7181,0.7050)(0.7223,0.7014)(0.7265,0.6974)
(0.7307,0.6933)(0.7349,0.6888)(0.7391,0.6848)(0.7433,0.6794)(0.7475,0.6745)(0.7517,0.6691)
(0.7559,0.6637)(0.7601,0.6578)(0.7643,0.6515)(0.7685,0.6452)(0.7727,0.6385)(0.7768,0.6317)
(0.7810,0.6241)(0.7852,0.6165)(0.7894,0.6088)(0.7936,0.6003)(0.7978,0.5917)(0.8020,0.5832)
(0.8062,0.5742)(0.8104,0.5652)(0.8146,0.5562)(0.8188,0.5468)(0.8230,0.5369)(0.8272,0.5274)
(0.8314,0.5175)(0.8356,0.5076)(0.8398,0.4973)(0.8440,0.4865)(0.8482,0.4748)(0.8523,0.4631)
(0.8565,0.4505)(0.8607,0.4371)(0.8649,0.4236)(0.8691,0.4083)(0.8733,0.3934)(0.8775,0.3777)
(0.8817,0.3620)(0.8859,0.3453)(0.8901,0.3278)(0.8943,0.3103)(0.8985,0.2918)(0.9027,0.2729)
(0.9069,0.2549)(0.9111,0.2361)(0.9153,0.2176)(0.9195,0.1996)(0.9237,0.1821)(0.9279,0.1655)
(0.9320,0.1484)(0.9362,0.1326)(0.9404,0.1174)(0.9446,0.1021)(0.9488,0.0881)(0.9530,0.0755)
(0.9572,0.0634)(0.9614,0.0531)(0.9656,0.0427)(0.9698,0.0342)(0.9740,0.0270)(0.9782,0.0202)
(0.9824,0.0148)(0.9866,0.0099)(0.9908,0.0063)(0.9950,0.0045)(0.9992,0.0048)};
\addlegendentry{all classes (0.81 at 0.258)}
\addplot[only marks,mark=o,mark size=2pt,blue,thick] coordinates {(0.258,0.81)};
\draw[blue,densely dotted] (axis cs:0.258,0) -- (axis cs:0.258,0.81);
\end{axis}
\end{tikzpicture}
\end{minipage}\hfill
\begin{minipage}[c]{0.40\textwidth}\centering
\begin{tikzpicture}[x=1.07cm,y=1.07cm,font=\footnotesize]
\foreach \r/\c/\v in {0/0/0.89, 0/2/0.16, 1/1/0.67, 1/2/0.84, 2/0/0.11, 2/1/0.33}{
  \pgfmathsetmacro{\g}{100*\v}
  \pgfmathsetmacro{\tc}{\v>0.5 ? 1 : 0}
  \fill[black!\g] (\c,-\r) rectangle ++(1,-1);
  \ifdim \v pt>0.5pt
    \node[white] at (\c+0.5,-\r-0.5) {\v};
  \else
    \node at (\c+0.5,-\r-0.5) {\v};
  \fi
}
\draw[black!40] (0,0) grid[step=1] (3,-3);
\foreach \i/\t in {0/arrow,1/cone,2/backgr.}{
  \node[anchor=east] at (0,-\i-0.5) {\t};
  \node[anchor=north, rotate=0] at (\i+0.5,-3.05) {\t};
}
\node[anchor=south] at (1.5,0.05) {\textbf{True}};
\node[anchor=south, rotate=90] at (-1.0,-1.5) {\textbf{Predicted}};
\end{tikzpicture}
\end{minipage}
\caption{Validation-split evaluation. Left: $F_1$--confidence curves of the deployed checkpoint (thick: all classes, maximum 0.81 at 0.258). Right: column-normalized confusion matrix at confidence 0.25 / IoU 0.45; each column sums to one over the predicted classes.}
\label{fig:eval}
\end{figure}

Table~\ref{tab:metrics} summarizes the detector. On the validation split the checkpoint reaches mAP@0.5 of 0.826 (arrows 0.94, cones 0.71); at the best-F1 threshold 0.258 ($F_1=0.81$) precision/recall are 0.92/0.89 for arrows and 0.81/0.65 for cones. The normalized confusion matrix (Fig.~\ref{fig:eval}, Ultralytics defaults: confidence 0.25, IoU 0.45) shows no arrow--cone confusion; residual errors are missed detections and background false positives. On the held-out test split mAP@0.5 is 0.720 (arrows 0.74, cones 0.70). Threshold and checkpoint were selected on the validation split, so the test figures provide a held-out evaluation; the arrow gap reflects the shift from web-sourced to field imagery and the small number of test arrows (541). The lower single-frame cone recall is mitigated operationally by the 30~Hz stream and the ten-frame filter of the BT. Exported to TensorRT INT8 (2.1~MB), reported latency on the Jetson Orin NX is 22~ms for inference (approximately 45 inferences/s) and 25~ms including pre/post-processing (approximately 40 frames/s); the reported power is 6.8~W.

\begin{table}[t]
\caption{Perception metrics of the deployed model. Navigation outcomes are reported per mission in Table~\ref{tab:missions}.}
\label{tab:metrics}
\centering\scriptsize
\setlength{\tabcolsep}{3pt}
\begin{tabular}{L{3.6cm}L{3.7cm}L{4.2cm}}
\toprule
\textbf{Metric} & \textbf{Value} & \textbf{Scope} \\
\midrule
mAP@0.5 / mAP@0.5:0.95 & 0.826 / 0.612 & validation split \\
mAP@0.5 / mAP@0.5:0.95 & 0.720 / 0.532 & held-out test split \\
Arrow AP@0.5, cone AP@0.5 & 0.94, 0.71 / 0.74, 0.70 & validation / test \\
Arrow P / R, cone P / R & 0.92 / 0.89, 0.81 / 0.65 & validation, conf.\ 0.258 \\
Arrow-direction accuracy & not independently established & no verified aggregate reported \\
Range / bearing accuracy & not established over full operating range & near-range jitter test only \\
Inference / pipeline latency & 22 / 25~ms & Jetson Orin NX, TensorRT INT8 \\
\bottomrule
\end{tabular}
\end{table}

\subsubsection{Arrow direction, distance and bearing.}
Arrow direction (left/right) is decided on the detector's box (Fig.~\ref{fig:photos}(a)) by majority vote of three geometric cues on the binarized arrow: mass asymmetry about the centroid, width gradient along the principal axis, and convex-hull sharpness at the extreme points. A verified aggregate accuracy and sample count for direction classification are not available; no numerical direction-accuracy claim is made. Distance is read from calibration tables that map box height (and width) to range, specified at 1~m stations from 1 to 10~m and interpolated piecewise-linearly; separate tables exist for the plate and the cone. A separate static laboratory sequence of 198 frames at 0.8--0.95~m showed box jitter of $\pm1.8$~px. This near-range observation does not establish range accuracy over 1--10~m, and an independent full-range field check was not performed. Bearing is computed as
\begin{equation}
\theta=\arctan\!\left(\frac{x_{\mathrm{box}}-c_x}{f_x}\right),\qquad c_x=320~\mathrm{px},\quad f_x=920~\mathrm{px},
\end{equation}
where $x_{\mathrm{box}}$ is the horizontal box-centre coordinate in pixels and $\theta$ is in radians. No independently verified bearing-error bound is reported.

\subsubsection{Classical baseline.}
To quantify the benefit of the learned detector on the same task we implemented a hand-crafted OpenCV detector representative of pre-CNN marker detection: arrows as dark, moderately non-convex polygonal blobs (5--10 vertices, solidity 0.5--0.95) inside segmented white plates; cones as saturated color blobs with vertical merging of stripe-split fragments and a taper test. Both detectors were scored on the 478-image test split with one greedy IoU-0.5 matcher (Table~\ref{tab:baseline}); the YOLOv8n numbers therefore differ slightly from the Ultralytics evaluator. Baseline AP@0.5 is 0.23 for arrows and 0.05 for cones, versus 0.68 for both with YOLOv8n: fixed color/shape rules fail across lighting, cone variants and background structures. The baseline code is in the released repository.

\begin{table}[t]
\caption{Learned detector vs.\ hand-crafted OpenCV baseline on the same held-out test split (478 images), scored using one greedy IoU-0.5 matcher, with P/R at each detector's best-$F_1$ point. The same checkpoint and images yield different AP in Table~\ref{tab:metrics} because that table uses the Ultralytics evaluator rather than this common matcher.}
\label{tab:baseline}
\centering\scriptsize
\setlength{\tabcolsep}{6pt}
\begin{tabular}{llccc}
\toprule
\textbf{Detector} & \textbf{Class} & \textbf{AP@0.5} & \textbf{Precision} & \textbf{Recall} \\
\midrule
YOLOv8n (deployed) & arrow & 0.68 & 0.89 & 0.62 \\
                   & cone  & 0.68 & 0.83 & 0.63 \\
Classical CV       & arrow & 0.23 & 0.56 & 0.33 \\
                   & cone  & 0.05 & 0.24 & 0.12 \\
\bottomrule
\end{tabular}
\end{table}

\section{Simulation-Based Integration}
\label{sec:sim}

Before field deployment, the stack was integrated in Gazebo with the rover URDF, steering kinematics and simulated depth, inertial and odometric inputs. The same EKF, RTAB-Map, Smac Hybrid, BT and P-controller plugin were used to resolve integration errors and tune the configuration. Continuous trajectory logs were not archived; consequently, no numerical comparison with MPPI, cross-track RMSE or terminal-error measurement is reported. A repeated instrumented controller benchmark is future work.

\section{Field Validation}
\label{sec:field}

\subsection{Platform}
\begin{figure}[t]
\centering
\begin{minipage}[b]{0.36\textwidth}\centering
\includegraphics[width=\linewidth]{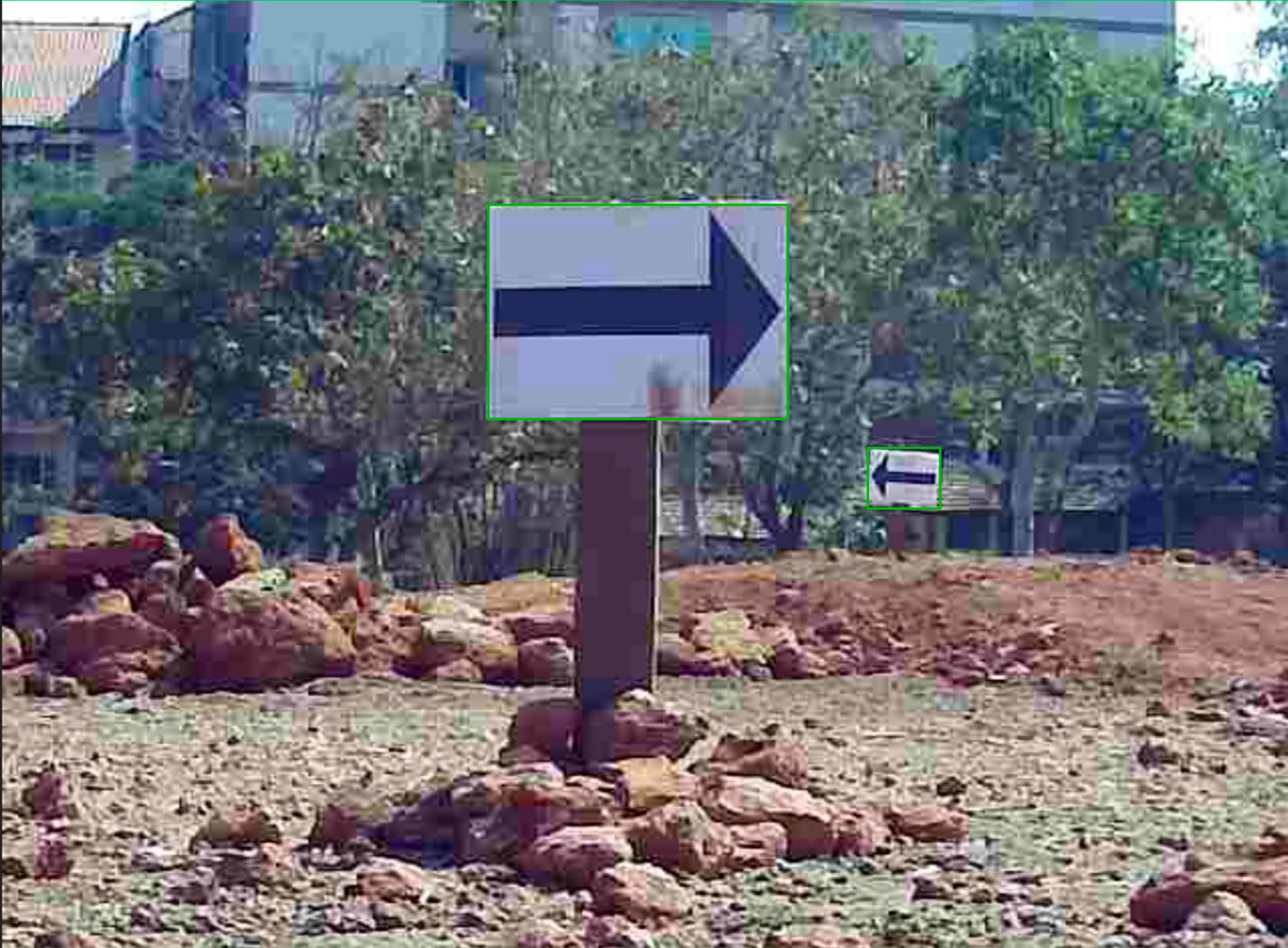}\\[-1pt]{\footnotesize (a)}
\end{minipage}\hfill
\begin{minipage}[b]{0.16\textwidth}\centering
\includegraphics[width=\linewidth]{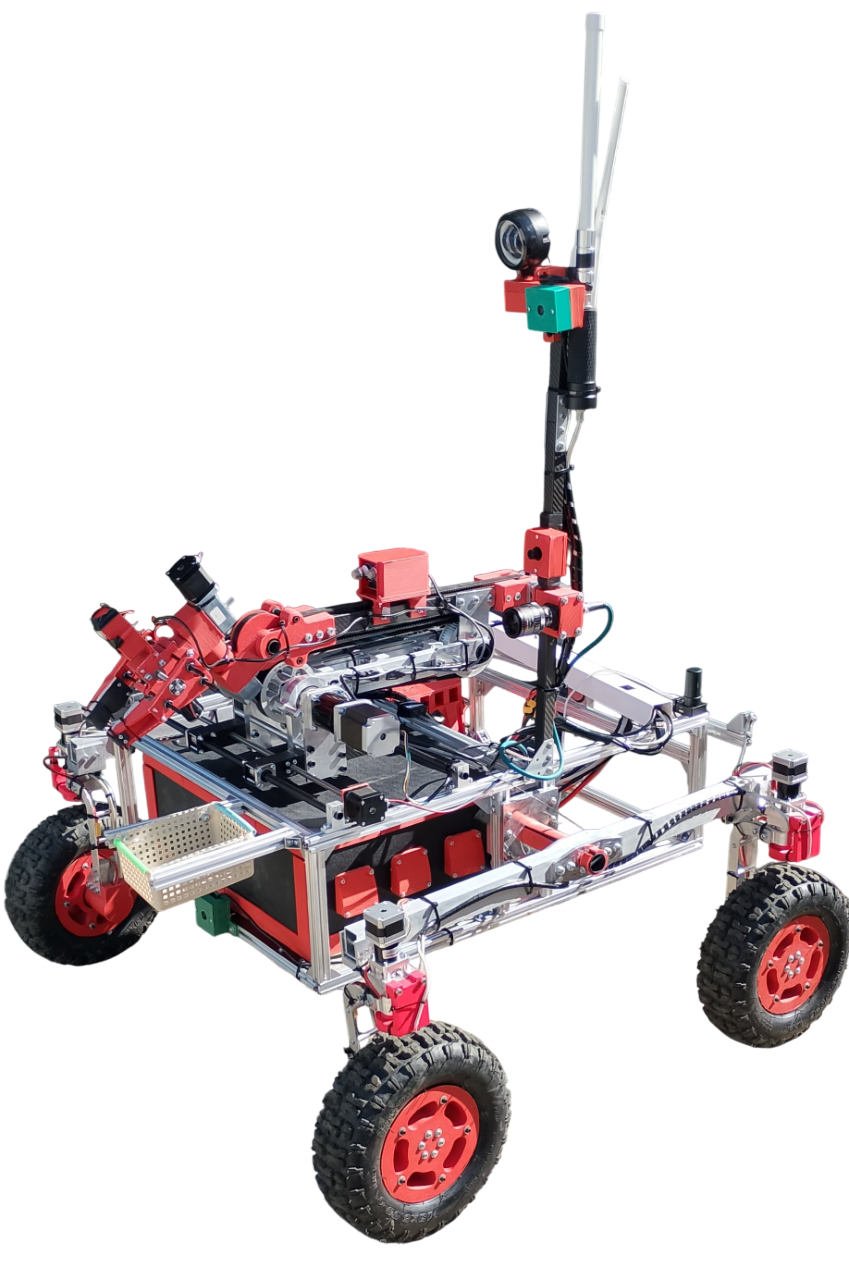}\\[-1pt]{\footnotesize (b)}
\end{minipage}\hfill
\begin{minipage}[b]{0.22\textwidth}\centering
\includegraphics[width=\linewidth]{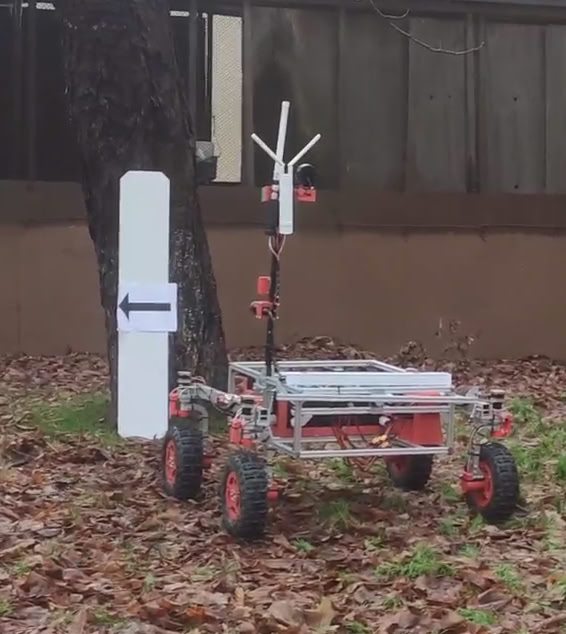}\\[-1pt]{\footnotesize (c)}
\end{minipage}
\caption{(a) Field detection with class, distance and bearing overlay; (b) the rover in field configuration; (c) final outdoor campaign on unprepared terrain.}
\label{fig:photos}
\end{figure}

The testbed ``Lancer Evo 1'' (Fig.~\ref{fig:photos}(b)) has a four-wheel rocker suspension with independently steered wheels, measures 105$\times$105$\times$53~cm, has 28~cm ground clearance, weighs 30~kg and carries up to 20~kg. Perception runs on an NVIDIA Jetson Orin NX, navigation on a Beelink SER5 Max mini-PC (AMD Ryzen~5 5500U); the two are linked by Ethernet under one ROS~2 Humble domain. Besides the Lucid camera and the D435i, the rover carries a WitMotion WT901BLECL IMU (added after the D435i's internal IMU proved too noisy for heading) and monitoring cameras. A \texttt{rosbridge}/\texttt{web\_video\_server} console shows video and telemetry; the operator issues no motion commands during autonomous runs.

\subsection{Task and Success Criteria}
The mission is autonomous navigation over moderately rugged terrain along a route of visual markers: black arrows on white plates point toward the next marker and an orange cone marks the terminal target. For each arrow the rover must detect it, estimate distance and bearing, generate a goal and enter a stop zone defined by the project specification as a circle of radius 2~m around the marker, keeping at least half of the rover inside it for at least 10~s; it then searches in the indicated direction. After reaching the cone it must return autonomously to the start. Collisions with markers or obstacles are not permitted. The project specification additionally lists a 5\% localization-error requirement; it was assessed by the test commission only qualitatively (coincidence of start and return coordinates on the operator console), so no numerical localization error is claimed in this paper.

\subsection{Campaigns and Per-Mission Results}
The test protocols document three complete autonomous missions in two two-hour campaigns~\cite{projectreport} (Table~\ref{tab:missions}); four hours is session time, not autonomous driving time. The preliminary indoor mission M1 (25~November 2025) was run in a test hall with above-average dust and an ambient temperature above 25~$^\circ$C, with markers placed within 3~m. It exposed the depth-range failure of Sect.~\ref{sec:nav} (space beyond about 3~m rasterized as occupied), which was fixed by the range filter and revalidated in the final outdoor campaign (4~December 2025) at the testing ground of the Research Institute of Mechanics, MSU: an unprepared area with uneven ground, tall grass and fallen leaves, light rain during the tests, and no artificial obstacles; the markers themselves and the natural terrain were the only obstacles. M2 and M3 used different start points and marker layouts, so no trajectory was replayed. In M3 the archived goal log\footnote{\url{https://msuroverteam.org/wp-content/uploads/2025/12/track2.txt}} contains three distinct goals; the straight segments between start, goals and return measure about 17~m, and the second leg placed the target about 10~m from the preceding stop. The protocols report recognition while moving, stops at the markers, cone arrival, autonomous return and no collisions; no recovery behaviors were reported. Return-to-start was accepted by the Nav2 goal checker, i.e.\ within 0.4~m and 0.5~rad of the stored start pose in the map frame; the residual error with respect to the physical start mark was not measured with an external reference. Console start/return coordinates coincided qualitatively~\cite{projectreport}. Route lengths of M1 and M2 and verified autonomous durations of all three missions are unavailable. The goal log does not independently establish a verified marker count for each mission. Stop-zone compliance was qualitative; the 0.4~m Nav2 tolerance is an acceptance setting, not a measured physical return error. A video of the final campaign is available online\footnote{\url{https://drive.google.com/file/d/11rRBgScfbcjOUezoQiBePUHVKEKXSEK9/view?usp=sharing}}.

\begin{table}[t]
\caption{Per-mission summary (``n/r'' = not recorded). The checklist score (Table~\ref{tab:acceptance}) is listed apart from the navigation outcomes. M1 ran before the depth-range fix (Sect.~\ref{sec:field}).}
\label{tab:missions}
\centering\scriptsize
\setlength{\tabcolsep}{4pt}
\begin{tabular}{L{3.6cm}L{2.4cm}L{2.5cm}L{2.4cm}}
\toprule
 & \textbf{M1 (indoor)} & \textbf{M2 (outdoor)} & \textbf{M3 (outdoor)} \\
\midrule
Date (session length) & 25 Nov 2025 (2~h) & 4 Dec 2025 (2~h, M2+M3) & 4 Dec 2025 \\
Conditions & indoor, dust, $>$25~$^\circ$C & grass, leaves, drizzle & grass, leaves, drizzle \\
Max.\ marker distance & $\leq$3~m & up to 10~m & $\approx$10~m \\
Route length (goal polyline) & n/r & n/r & $\approx$17~m \\
Markers recognized while moving / stop zones held / cone reached (per protocol) & all / all / yes & all / all / yes & all / all / yes \\
Return to start (within 0.4~m goal tolerance) / error vs.\ ground mark & yes / n/r & yes / n/r & yes / n/r \\
Collisions / recovery behaviors & 0 / 0 & 0 / 0 & 0 / 0 \\
Checklist score (Table~\ref{tab:acceptance}) & 96\% & 99\% & 100\% \\
\bottomrule
\end{tabular}
\end{table}

\begin{table}[t]
\caption{Specification-based acceptance checklist (weights sum to 100). Deductions concern only the reusable route map: not exercised in M1, generated but not saved in M2.}
\label{tab:acceptance}
\centering\scriptsize
\setlength{\tabcolsep}{5pt}
\begin{tabular}{lcccc}
\toprule
\textbf{Criterion} & \textbf{Weight} & \textbf{M1} & \textbf{M2} & \textbf{M3} \\
\midrule
Recognition of all visual markers        & 20  & 20 & 20 & 20  \\
Autonomous navigation, return, reusable map & 20  & 16 & 19 & 20  \\
Low-level motion control                 & 20  & 20 & 20 & 20  \\
Hardware reliability                     & 20  & 20 & 20 & 20  \\
Operator telemetry and video console     & 6   & 6  & 6  & 6   \\
Real-time integration of all components  & 10  & 10 & 10 & 10  \\
Stop-and-document at each marker         & 4   & 4  & 4  & 4   \\
\midrule
Total                                    & 100 & 96 & 99 & 100 \\
\bottomrule
\end{tabular}
\end{table}

The checklist (Table~\ref{tab:acceptance}) is a weighted specification-compliance score, not navigation accuracy. Separate development-stage capability tests are excluded from the mission count. Three missions demonstrate integration, not statistical reliability.

\section{Conclusions}

The released stack integrates vision, RGB-D SLAM, Ackermann-feasible planning, proportional path following and behavior-tree mission management with rover actuation. Three documented missions completed marker-guided navigation and return to start without operator motion commands. Limitations include the small mission count, unmeasured physical return error and full-range perception accuracy, an uninstrumented controller comparison, unchecked search-rotation swept footprints and portability not yet demonstrated on a second platform. Future work targets these limitations through repeated instrumented trials.

\section*{Disclosure of Interests}
The authors have no competing interests to declare that are relevant to the content of this article.

\end{document}